# THE BARRIER OF MEANING IN ARCHAEOLOGICAL DATA SCIENCE

Luca Casini*, Marco Roccetti*, Giovanni Delnevo*, Nicolò Marchetti**, Valentina Orrù**

*Alma Mater Studiorum - University of Bologna
Department of Computer Science and Engineering
E-mail: {luca.casini7; marco.roccetti; giovanni.delnevo2}@unibo.it

**Alma Mater Studiorum – University of Bologna
Department of History and Cultures
E-mail: {nicolo.marchetti; valentina.orru}@unibo.it

## KEYWORDS

Human intelligence, Deep Learning, Archaeology, Data Science.

## ABSTRACT

Archaeologists, like other scientists, are experiencing a data-flood in their discipline, fueled by a surge in computing power and devices that enable the creation, collection, storage and transfer of an increasingly complex (and large) amount of data, such as remotely sensed imagery from a multitude of sources. In this paper, we pose the preliminary question if this increasing availability of information actually needs new computerized techniques, and Artificial Intelligence methods, to make new and deeper understanding into archaeological problems. Simply said, while it is a fact that Deep Learning (DL) has become prevalent as a type of machine learning design inspired by the way humans learn, and utilized to perform automatic actions people might describe as intelligent, we want to anticipate, here, a discussion around the subject whether machines, trained following this procedure, can extrapolate, from archaeological data, concepts and meaning in the same way that humans would do. Even prior to getting to technical results, we will start our reflection with a very basic concept: Is a collection of satellite images with notable archaeological sites informative enough to instruct a DL machine to discover new archaeological sites, as well as other potential locations of interest? Further, what if similar results could be reached with less intelligent machines that learn by having people manually program them with rules? Finally: If with barrier of meaning we refer to the extent to which human-like understanding can be achieved by a machine, where should be posed that barrier in the archaeological data science?

## INTRODUCTION

Artificial Intelligence (AI) is a field much older than most people would suspect, with an history that dates back to the 1940s, and that has lived through multiple rises and falls in popularity over the decades. However, in the last few years it has seen an incredible renaissance, and applications of it found their way in the everyday life of billions of people. What once was a word from the realm of science fiction has now become a tangible part of our lives. The impact of AI on society is unprecedented and there seems to be no field where its potential is not applicable.

It may be surprising, yet the technology behind this new AI revolution, as the field itself, is also not that new. The engine under the hood of modern AI is a special kind of probabilistic model that has its roots in the late 1950s, called Artificial Neural Network (ANN), that is able to learn even very complex relations hidden in the data, provided a sufficient amount of training examples (Rosenblatt 1958).

The name comes from the fact that these models are vaguely inspired by our neurons; they are comprised of interconnected units that get a stimulus signal as input and perform a simple operation that is cast to all the other connected units, downstream. In their early years, ANNs were difficult to train, due to the lack of good algorithms, sufficient computer power and enough data, eventually going out of fashion. But as the 21$^{st}$ century approached, new algorithms were developed, compute power grew exponentially and, most importantly, the Internet brought on the so-called *big data*: the perfect fuel for ANNs. With these new resources, researchers started building bigger and bigger models with many layers of stacked artificial neurons, leading to the birth of a new research field called Deep Learning (DL).

Image analysis and classification were among the first areas that showed the great potential of DL, with the rise of Convolutional Neural Networks (CNN), a special neural architecture inspired by the human visual cortex that emerged in the 90s (LeCun et al. 1989). CNNs revolutionized image processing, by surpassing state of the art methods, around 2012 (Krizhevsky et al. 2012; Ciresan et al. 2011), and then soon after competing even with human performances in many classification and recognition tasks (Esteva et al. 2017; Nam et al. 2018; He et al. 2015).

As of now, they are the staple of any attempt to automate a task that involves images of any kind, yet an important consideration should be made: DL models excel at tasks that involve perception, but perception is only one aspect of intelligence. What we perceive, in turn, informs our internal representation of the world, and this representation is the basis upon which we build our reasoning, finally leading us to decisions and actions in the real world.

Thus, when dealing with these technological objects one should not forget that their inner workings is somewhat limited to the perception sphere and that their output (i.e., their decisions) is only informed by this perception, and not by any ulterior knowledge that, instead, a human expert may already possess or abstract by the same information.

Inspired by this kind of studies (Mitchell 2009; Lalumera et al. 2019), we refer to this difference with the term of *barrier of meaning*.

Along this line of reasoning, this paper challenges that concept of barrier of meaning in the context of archaeological data (Marchetti et al 2018). The problem being whether a DL model, trained on imagery from a multitude of sources, is able to recognize new locations of interest, better than a combination of human experts and traditional programs instructed with rules would do. To this aim, in the next Section, we will discuss our archaeological case study, while the concept of barrier of meaning is debated in the subsequent Section. The final Section concludes the paper.

**CASE STUDY: ARCHAEOLOGICAL DATA SCIENCE**

To illustrate our point, in this Section we will introduce a archaeological data science case study, with the aim of automatically spotting certain areas of interest from satellite images. As any other discipline in the recent years, archaeology too has experienced the impact of big data coming from the innovations in the field of remote sensing (the act of capturing data remotely, e.g., with satellites); like never before there is the possibility of working with high resolution aerial and satellite images from any part of the world, captured with a plethora of techniques, that go from simple optical sensors to more peculiar ones, like SAR or LIDAR (that may help highlight certain physical phenomena).

Improved is also the speed at which this data is produced: the European Space Agency's Sentinel-1 satellites can scan the entire planet every 6 days, thus opening new possibilities for monitoring how a scenery evolves over time, and after certain events. In the context of archaeology, these new resources allow researchers to pinpoint interesting sites without the need to leave their offices, by looking for certain archaeological features that appear on satellite imagery. This remote sensing phase is very important in the preparation of a mission, because it can save time and resources, by identifying the most promising sites to visit before leaving for the field (Hritz, 2010).

The downside is that it requires an experienced archaeologist to thoroughly inspect every corner of a map that can easily span thousands of square kilometers, personally searching for what looks like one of the areas of interest. Given the sparsity of such sites, whoever is tasked with this job will probably and unfortunately spend most of the time looking at *empty* places; further, it should be considered that the fallibility of humans also means that some site will be missed or misclassified.

In our specific case, experienced archaeologists investigated an area (visible in Figure 1): about 2000 square kilometers wide, in the Iraqi region of Qadisiyah. Inside this area, 166 polygons were individuated (in addition to the 415 registered by Adams, 1981), representing the extent of dig sites that were: i) initially identified with remote sensing, and later ii) visited and verified on the field. It is worth noticing that 21 of those 166 sites were actually misidentified as sites at the beginning, and then recognized as non-sites once visited. Examples of those sites/non-sites are shown in Figures 2 and 3 (Marchetti 2018; Marchetti et al. 2017).

A first consideration amounts to the fact that the dimension of all those 166 polygons varies considerably, going from as little as 300 square meters to as much as 1.5 square kilometers. 19 optical images, with a digital size of 23000x23000 pixels and a resolution in the order of few meters, cover that investigated area (including all the 166 sites). Considered could be also additional contextual information, in the form of CORONA photos, taken during the 1960s by USA spy satellites (Pournelle, 2007). Those photos have the advantage of being taken in a timeframe when the human impact on the area was minimal, but also have the downside of a considerably lower resolution, as well as the presence of cloud-like obfuscation.

Within this context, in the next Subsections, we will discuss on a proposal, based on DL, aimed at differentiating automatically sites from non-sites; this also in contrast with a simpler and more traditional approach based on a rule-based system that would not need to learn anything.

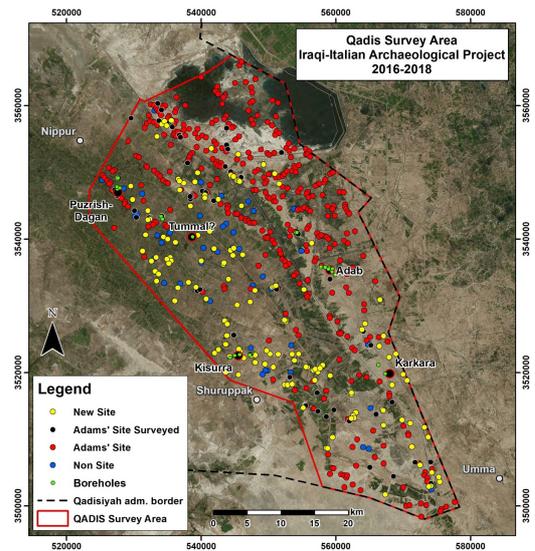

Figure 1. The great area to be investigated

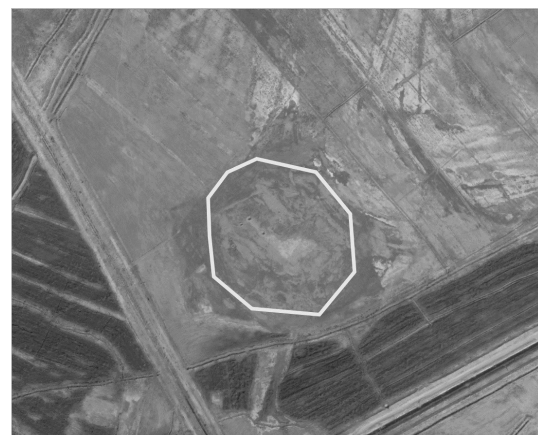

Figure 2. A *site* example

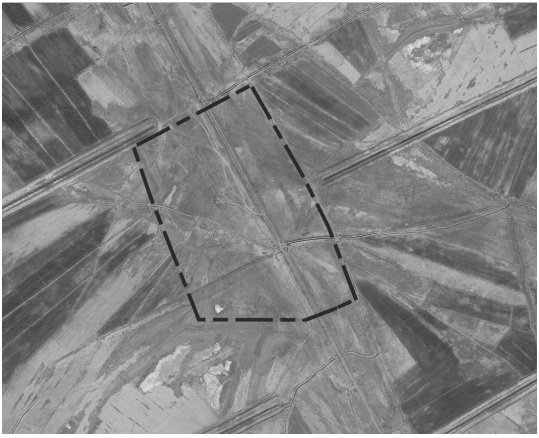

Figure 3. A *non-site* example

**A DL Approach**

The problem of finding areas of interest from satellite images has already emerged in a variety of different fields (beyond archaeology), with many proposed DL solutions, although all of them differ in the characteristics of the final goal and, more importantly, of the type of the dataset on which learning algorithms should be trained.

The first thing to notice about the present dataset is its peculiar size, being both small and big at the same time; in the sense that we have a very *few* labeled examples with images that are instead quite *unwieldy*.

Let us focus on the first part. The scarcity of labeled examples is one of the key obstacles to overcome for deep learning to be used, but there are few techniques one could try. For example, *data augmentation*. It consists in taking some few labeled examples to which apply some transformation rules (e.g., translations, rotations, mirroring and shearing) that extend the initial dataset, without changing the general meaning, (Cheng et al. 2016). Another viable alternative to extend the dataset amounts to *transfer learning* (Shin et al. 2016). Here the idea is that of using a DL model, already trained in a similar task, and to adapt it to a new situation, while retaining all the useful knowledge it has already learned. What one would do here is to take some popular pre-trained DL model for image recognition, like ResNet (He et al. 2016) or Inception (Szegedy et al. 2015), and replace the last layers, responsible for classification, with the new ones tailored for the archaeological task. In essence, we would leverage on the pre-trained model for the feature extraction part (the one that usually requires more training examples), while we would leave the conclusion of the training activity to the few specific examples of site/non-sites that we possess. Nonetheless, the real utility of this approach in this case is questionable, depending on the fact that the aforementioned pre-trained models typically include images which are very different, in nature, from those we are interested in. There is also a possibility worth exploring, namely the so called *semi-supervised learning*. What we have described so far, in fact, is a part of a family of learning algorithms called *supervised*, as they need labeled examples to be trained. On the other hand, unsupervised models do not need labels, and learn by differentiating data directly, just *looking at* them. Semi-supervised techniques fall in-between, where one uses the few labels available to extrapolate spare labels of data, that are either close or similar to those that are available (W. Han et al. 2018; Dópido et al. 2013; J. Han et al. 2015).

Said about the trouble of the scarcity of training examples, we can move to another problem that characterizes this study. In fact, if on one side the scarcity of labeled examples is an issue, on the other side, equally crucial is the problem of managing very large images. Indeed, 19 photos, each of a size of 23000x23000 pixels, represent an enormous area to scan, both for a human inspection and also if a CNN is asked to do that, in a single step. A solution to this problem would be obviously to divide the area into tiles, and then train a classifier on those tiles, considering each of them separately. Doing this, one would end up with around 190,000 tiles, each of size 230x230 pixels, that is with a reasonable extension of typical images managed by CNNs. Here, the problem is that most of those tiles would come out *empty* (without relevant archaeological information), and if one were to consider all those empty as negative examples, s/he would end up with a very imbalanced dataset, in favor of negative examples (Kotsiantis et al. 2006). Indeed, the real negative examples of this case are just those 21, of which the archaeologists have checked the *negativity* on the field. What one could do then is to create a larger *negative class* including both the 21 locations certified as *non-sites* plus a random sample from tiles that cannot be labeled as *positive*. As a natural consequence of this solution, to better tune the DL model, it would be mandatory to differentiate among them (real negative vs. empty) during the testing phase, in order to see if and how the model makes mistakes in their classification.

The *positive* class, the *sites,* also poses some questions. Many of those sites are large in geospatial dimensions, and would not fit inside a single tile. Breaking down a site into more than one tile, and considering them all *positives*, looks like an acceptable proposal. Yet, in this case this choice leaves open the possibility of having a tile marked as positive, while actually not containing anything useful. Either because the part of the polygon inside is too small to give information, or because that part is not informative. A solution is to discard those tiles whose informative content is too scarce to represent positive examples. An alternative would be also that of engaging an archaeologist in the fine activity of assigning a *meaning* only to those specific parts of the polygon that represent relevant information, and then center the tile around them. Lastly, there is uncertainty on how to treat cases in which *site* and *non-site* instances are present in the same tile; this circumstance could be quite rare (at least in our dataset), but it is interesting, nonetheless from a general standpoint. A hypothesis would be to treat them as positive examples, as not doing so would generate a *false negative* error, while the opposite would just represent a case of a *noisy* positive example. To conclude, at this stage of the present project, we do not have yet answers to all the matters presented here, the aim of this article being that of emphasizing how issues that look like subtleties are actually very important to the success/failure of this project.

**Do We Really Need DL?**

Given the popularity of DL in the present days, one is often tempted to adopt it as a solution, even if it is not necessary at all, often introducing a series of problems specific to DL that would be avoided otherwise: namely, the fact that ANNs are black boxes that involve chance and uncertainty, as well as their need for considerable amounts of data. Discussing this project, instead, we asked ourselves if a DL-based approach were truly necessary, and what other option we had to create a solution that did not involve DL. Not only. Even if more traditional computerized methods were not that able in recognizing sites of archaeological interest, they would still represent a good baseline against which we could contrast more complex DL models. Along this line of reasoning, for example, one could think of an algorithm that uses classical computer vision techniques, and looks for a certain type of shape and color in a map, to find a site of interest. Obviously shapes and colors should be predefined in this case, with the help of the archaeologists. Nonetheless, such similar techniques yield, on average, accuracy results, whose validity is questionable and mostly depending on the very specific context of application (Jain et al. 1998; Lefevre et al. 2007; Liu et al. 2013). In addition, to cite is also the fact that the feature extraction activity mentioned above would require a lot of work, that is hardly automatable. From this perspective, if the goal is to save time by not looking personally at a map (while using this time for other tasks), the adoption of classical computer vision techniques does not look like a really smart idea. Obviously, learning features automatically would help, like in (Caspari et al. 2014; Gall et al. 2011; Lowe 2004), but then one would slip again into the machine learning territory, with all the implications we have already discussed above.

**THE BARRIER OF MEANING**

During the discussion of this problem, it has emerged that many DL design choices are bound to the interpretations of the context that only a domain expert (i.e., an archaeologist) and a data scientist together can give. This brings us to the concept of so called *barrier of meaning* (Mitchell, 2009). What we want to define, by adopting this term, is the divide between the knowledge in the expert's mind and the knowledge grasped by the machine; in other words, the gap between the real phenomenon/process we are trying to automate with a CNN and the actual approximation of it that gets learnt at the end. While designing a DL-based system, one should not forget the existence of this gap, simply because it is of great relevance to remember that the decisions that a DL machine takes directly depend on its input and on what can be abstracted from it, without any additional information being available. For humans (and experts of a given field), the game is just different fortunately, as they have additional contextual knowledge, experiences and comprehension of real phenomena that inform their choices and that the machine simply does not have. If this additional context can be somehow encoded as an input for the machine, then there is a chance that it may help, but oftentimes this encoding is not possible as even the human experts are not perfectly conscious of how they formulate decisions (Delnevo et al. 2019). Consequently, once the machine is trained, one should not forget the way it makes decisions and, more importantly, the way it makes mistakes are surely not the same as how a human (expert) would do. Especially mistakes make this very clear, and there are countless examples of DL misclassifying objects, that even a child would get correctly (Ribeiro et al. 2016). Another aspect, even more important to consider, is that the knowledge of an expert is not always neutral, nor a guarantee of truthiness. For example, in the field of medical image diagnosis, consensus among experts is an open issue, as there is not a definitive standard on how to interpret certain situations (Lalumera et al. 2019). Even the dataset that is used to train a DL model is subjected to this phenomenon, and, as such, this cannot be considered a precise description of the reality, rather just a representation one's beliefs/methodology. In conclusion, DL models are just as reliable as the data they were trained on. When we see models with good prediction performances, especially in crucial fields like in medical diagnostics, we must avoid considering them as something *superhuman* (that is, *more intelligent* than a human), rather as something that is very good at perceiving/recognizing very specific features yet, at the same time, very limited in reasoning (Geirhos et al. 2018). Putting humans, both experts and users, at the center of the AI design, in a sort of human-centered machine learning loop (Gillies et al. 2016; Palazzi et al. 2004; Roccetti et al. 2010), may help trespass the barrier of meaning, or at least may represent a step in the right direction. For example, some concrete actions, along this path, would be those of deciding what metrics to maximize and how to treat errors (Roccetti et al. 2019); also in the case of archaeology, in fact, relevant is whether the archaeologists prefer to avoid either false positives (as they are a waste of time) or false negatives (missing even a single site could be a crucial issue) or both (to reach a sort of equilibrium).

**CONCLUSIONS**

We presented a case study of archaeological data science we are about to work on, as a proxy to describe the criticalities that arise when working with DL in crucial human-centric contexts. We proposed a possible path to follow for this research, and also highlighted some of the more crucial implications, along with possible viable solutions. As a part of this discussion, we focused on the concept of barrier of meaning. Being this concept crucial to understand the extent of the divide between how knowledge is conceptualized and elaborated in an expert's mind, as opposed to the way the machine *operates*.
We posit that neglecting this issue during the very first design phase of a DL model can result in both a scarce precision of the model, as well as in frequent misunderstandings among experts of different fields. Not only, if experts are not made aware of these aspects, this may lead them to believe that a model is more *intelligent* than it really is, and, in turn, they will fail to recognize that errors committed by the machine are of a different nature than those a human would commit.